%% file: main.tex
\documentclass{article}

    \usepackage[preprint]{neurips_2021}

\usepackage[utf8]{inputenc} % allow utf-8 input
\usepackage[T1]{fontenc}    % use 8-bit T1 fonts
\usepackage{hyperref}       % hyperlinks
\usepackage{url}            % simple URL typesetting
\usepackage{booktabs}       % professional-quality tables
\usepackage{amsfonts}       % blackboard math symbols
\usepackage{nicefrac}       % compact symbols for 1/2, etc.
\usepackage{microtype}      % microtypography
\usepackage{xcolor}         % colors

\usepackage{algorithm}     
\usepackage{algorithmic}

\usepackage{graphics} 
\usepackage{graphicx} 
\usepackage{wrapfig}

\usepackage{tikz}
\usetikzlibrary{bayesnet2}
\usepackage{subcaption}

\input{math_commands.tex}

\title{Training on Test Data with Bayesian Adaptation for Covariate Shift}

\author{%
  Aurick Zhou, Sergey Levine \\
  Department of Electrical Engineering and Computer Sciences\\
  University of California, Berkeley\\
  \texttt{\{aurick,svlevine\}@berkeley.edu} 
}

\begin{document}

\maketitle

\begin{abstract}
When faced with distribution shift at test time, deep neural networks often make inaccurate predictions with unreliable uncertainty estimates.
While improving the robustness of neural networks is one promising approach to mitigate this issue, an appealing alternate to robustifying networks against all possible test-time shifts is to instead directly adapt them to unlabeled inputs from the particular distribution shift we encounter at test time.
However, this poses a challenging question: in the standard Bayesian model for supervised learning, unlabeled inputs are conditionally independent of model parameters when the labels are unobserved, so what can unlabeled data tell us about the model parameters at test-time? 
In this paper, we derive a Bayesian model that provides for a well-defined relationship between unlabeled inputs under distributional shift and model parameters, and show how approximate inference in this model can be instantiated with a simple regularized entropy minimization procedure at test-time. 
We evaluate our method on a variety of distribution shifts for image classification, including image corruptions, natural distribution shifts, and domain adaptation settings, and show that our method improves both accuracy and uncertainty estimation.
\end{abstract}

\input{intro}

\input{method}

\input{related_work}

\input{experiments}

\input{conclusion}

%%%%%%%%%%%%%%%%%%%%%%%%%%%%%%%%%%%%%%%%%%%%%%%%%%%%%%%%%%%%
\bibliographystyle{abbrvnat} 
\bibliography{references}
% \newpage 
% \input{checklist}

\newpage
\appendix
\input{appendix}

\end{document}

%% file: math_commands.tex
\usepackage{amsmath,amsfonts,bm}

\def\eqref#1{equation~\ref{#1}}
\def\1{\bm{1}}

\DeclareMathAlphabet{\mathsfit}{\encodingdefault}{\sfdefault}{m}{sl}
\SetMathAlphabet{\mathsfit}{bold}{\encodingdefault}{\sfdefault}{bx}{n}

\newcommand{\E}{\mathbb{E}}

\DeclareMathOperator*{\argmax}{arg\,max}

%% file: intro.tex
\section{Introduction}
Modern deep learning methods can provide high accuracy in settings where the model is evaluated on data from the same distribution as the training set, but accuracy often degrades severely when there is a mismatch between the training and test distributions \citep{Hendrycks2019BenchmarkingPerturbations, Taori2020MeasuringClassification}. 
In safety-critical settings,
effectively deploying machine learning models requires not only high accuracy, but also requires the model to reliably quantify uncertainty in its predictions in order to assess risk and potentially abstain from making dangerous, unreliable predictions.
Reliably estimating uncertainty is especially important in settings with distribution shift where inaccurate predictions are more prevalent, but the reliability of uncertainty estimates often also degrades along with the accuracy as the shifts become more severe \citep{Ovadia2019CanShift}. 
In real-world applications, distribution mismatch at test time is often inevitable, thus necessitating methods that can robustly handle distribution shifts, both in terms of retaining high accuracy but also in providing meaningful uncertainty estimates.

As it can be difficult to train a single model to be robust to all potential distribution shifts we might encounter at test time, we can instead robustify models by allowing for \textit{adaptation} at test time, where we finetune the network on unlabeled inputs from the shifted target distribution, thus allowing the model to specialize in the particular shift it encounters.
Since test-time distribution shifts often cannot be anticipated during training time, we restrict the adaptation procedure to operate without any further access to the original training data.
Prior work on test-time adaptation \citep{Wang2020Tent:Minimization, Sun2019Test-TimeShifts} focused on improving accuracy, and found that simple objectives like entropy minimization capable of providing substantial improvements under distribution shift \citep{Wang2020Tent:Minimization}.
However, these prior works do not consider uncertainty estimation, and an objective like entropy minimization can quickly make predictions overly confident and less calibrated, leaving us without reliable uncertainty estimates and thus unable to quantify risks when making predictions.
Our goal in this paper is to design a test-time adaptation procedure that can not only improve predictive accuracy under distribution shift, but also provide reliable uncertainty estimates. 

While a number of prior papers have proposed various heuristic methods for test-time adaptation~\citep{Wang2020Tent:Minimization, Sun2019FunctionalNetworks}, it remains unclear what precisely unlabeled test data under covariate shift can actually tell us about the optimal classifier. 
In this work, we take a Bayesian approach to this question, and explicitly formulate a Bayesian model that describes how unlabeled test data from a different domain can be related to the classifier parameters. 
Such a model requires introducing an additional explicit assumption, as the classifier parameters are conditionally independent of unlabeled data in the standard model for discriminative classification \citep{Seeger2000Input-dependentModels}. 
The additional assumption we introduce intuitively states that the data generation process at test-time, though distinct from the one at training time (hence, under covariate shift) is still more likely to produce inputs that have a single unambiguous labeling, even if that labeling is not known. 
We argue that this assumption is reasonable in practice, and leads to an appealing graphical model where approximate inference corresponds to a Bayesian extension of entropy minimization.

We propose a practical test-time adaptation strategy, Bayesian Adapatation for Covariate Shift (BACS), which approximates Bayesian inference in this model and outperforms prior adaptive methods both in terms of increasing accuracy and improving calibration under distribution shift.
Our adaptation strategy is simple to implement, requires minimal changes to standard training procedures, and outperforms prior test-time adaptation techniques on a variety of benchmarks for robustness to distribution shifts.

%% file: method.tex
\newcommand{\x}{\mathbf{x}}
\newcommand{\X}{\mathbf{X}}

\section{Bayesian Adaptation for Covariate Shift}
In this section, we will devise a probabilistic graphical model that describes how unlabeled data in a new test domain can inform our posterior about the model, and then describe a practical deep learning algorithm that can instantiate this model in a system that enables test-time adaptation. 
We will begin by reviewing standard probabilistic models for supervised and discuss why such models are unable to utilize unlabeled data.
We then discuss a probabilistic model proposed by \citet{Seeger2000Input-dependentModels} that does incorporate unlabeled data in a semisupervised learning (SSL) setting, and propose an extension to the model to account for distribution shift. 
Finally, we discuss the challenges in performing exact inference in our proposed model and describe the approximations we introduce in order to derive a tractable inference procedure suitable for test-time adaptation.

\subsection{Probabilistic Model for Covariate Shift} 

\begin{wrapfigure}{r}{0pt}
      \begin{tikzpicture}[scale=0.9, transform shape]
      
      \vspace{-30pt}
      % Define nodes
      \node[obs]                               (x) {$\x$};
      \node[obs, right=of x]                               (y) {$y$};
      \node[latent, above=of y] (theta) {$\theta$};
      \node[latent, above=of x]  (phi) {$\phi$};
    
      % Connect the nodes
      \edge {x,theta} {y} ; %
      \edge {phi}     {x} ; % 
    
      % Plates
      \plate {yx} {(x)(y)} {$N$} ;
    \end{tikzpicture}
    \caption{\footnotesize Probabilistic model for standard supervised learning, observing $N$ labeled data points.}
    \vspace{-15pt}
    \label{fig:sl_gm}
\end{wrapfigure}
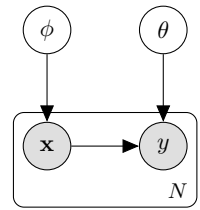
In the standard Bayesian model for supervised learning (Figure \ref{fig:sl_gm}), we assume that the inputs $\X$ are sampled i.i.d. from a generative model with parameters $\phi$, while the corresponding labels are sampled from a conditional distribution $p(Y\vert \X, \theta)$ parameterized by $\theta$. 
The parameters $\theta$ and $\phi$ are themselves random variables, sampled independently from prior distributions $p(\phi)$ and $p(\theta)$.
We observe only a dataset $\mathcal D = \{\X_i, Y_i\}_{i=1}^n$, and then perform inference over the parameters $\theta$ using Bayes rule:
\begin{align}
    p(\theta \vert \mathcal D) \propto p(\theta) \prod_{i=1}^n p(Y_i \vert \X_i, \theta).
\end{align}
To make predictions on a new input $\x_{n+1}$, we marginalize over the posterior distribution of the classifier parameters $\theta$ to obtain the predictive distribution
\begin{align*}
    p(Y_{n+1} \vert \X_{n+1}, \mathcal{D}) = \int p(Y_{n+1}\vert \X_{n+1}, \theta) p(\theta\vert \mathcal D) \, d\theta.
\end{align*}

Note that observing the (unlabeled) test input $\x_{n+1}$
(or more generally, any number of test inputs) does not affect the posterior distribution of parameters $\theta$, and so within this probabilistic model, there is no benefit to observing multiple unlabeled datapoints before making predictions.

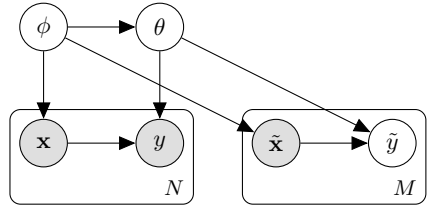
\begin{wrapfigure}{r}{0pt}
  \begin{tikzpicture}[scale=0.9, transform shape]

  % Define nodes
  \node[obs]                               (x) {$\x$};
  \node[obs, right=of x]                               (y) {$y$};
  
  \node[obs, right =of y]                               (tx) {$\tilde \x$};
  \node[latent, right=of tx]                               (ty) {$\tilde y$};
  \node[latent, above=of y] (theta) {$\theta$};
  \node[latent, above=of x]  (phi) {$\phi$};
  
  % Connect the nodes
  \edge {x,theta} {y} ; %
  \edge {phi}     {x} ; % 
  \edge{phi}      {theta}  % 
  
  \edge {tx,theta} {ty} ; %
  \edge {phi}     {tx} ; % 

  % Plates
  \plate {yx} {(x)(y)} {$N$} ;
  \plate {txy} {(tx)(ty)} {$M$} ;
\end{tikzpicture}
\caption{\footnotesize Model for SSL \citep{Seeger2000Input-dependentModels}, observing $N$ labeled pairs and $M$ unlabeled inputs, both generated from the same distribution. Observing unlabeled inputs $\tilde \x$ provides information about the input distribution $\phi$, and thus about $\theta$.}
\label{fig:ssl_gm}
\vspace{-20pt}
\end{wrapfigure}

\textbf{Inference for semi-supervised learning}.
By treating the parameters $\theta, \phi$ as a-priori independent, the standard model assumes there is no relationship between the input distribution and the process that generates labels, leading to the inability to utilize unlabeled data in inferring $\theta$. 
To utilize unlabeled data, we can introduce assumptions about the relationship between the labeling process and input distribution using a model of the form in Figure \ref{fig:ssl_gm} \citep{Seeger2000Input-dependentModels}.

The assumption we use has a simple intuitive interpretation: we assume that the inputs that will be shown to our classifier have an unambiguous and clear labeling. This assumption is reasonable in many cases: if the classifier discriminates between automobiles and bicycles, it is reasonable that it will be presented with images that contain either an automobile or bicycle. Of course, this assumption is not necessarily true in all settings, but this intuition often agrees with how discriminatively trained models are actually used. This simple intuition can be formalized in terms of a prior belief about the \emph{conditional entropy} of labels conditioned on the inputs.

Similar to \citet{Grandvalet2004Semi-supervisedMinimization}, we can encode this belief using the additional factor
\begin{align} \label{eq:ent-prior-ssl}
    p(\theta \vert \phi) &\propto \mu(\theta) \exp(-\alpha H_{\theta, \phi}(Y\vert \X)) \\
    &= \mu(\theta) \exp \left(\alpha \E_{\X \sim p(X\vert \phi), Y\sim p(Y\vert X, \theta)}[\log p(Y\vert \X, \theta)] \right),
\end{align}
where $\mu(\theta)$ is a prior over the parameters $\theta$ that is agnostic of the input distribution parameter $\phi$.

Now, observing unlabeled test inputs $\mathcal U = \{\tilde \x_1,\ldots, \tilde \x_m\}$ provides information about the parameter $\phi$ governing the input distribution, which then allows us to update our belief over the learned parameters $\theta$ through Equation \ref{eq:ent-prior-ssl} and thus allows inference to utilize unlabeled data within a Bayesian framework.

\textbf{Extension to covariate shift}. The previous probabilistic model for SSL assumes all inputs were drawn from the same distribution (given by $\phi$). However, our goal is use unlabeled data to adapt our model to a \textit{different} test distribution, so we extend the model to incorporate \textit{covariate shift}.

\begin{wrapfigure}{r}{0pt}
  \vspace{-5pt}
  \begin{tikzpicture} [scale=0.9, transform shape]

  % Define nodes
  \node[obs]                               (x) {$\x$};
  \node[obs, right=of x]                               (y) {$y$};
  
  \node[latent, right=of y]                               (ty) {$\tilde y$};
  \node[obs, right= of ty]                               (tx) {$\tilde \x$};
  \node[latent, above=of y] (theta) {$\theta$};
  \node[latent, above=of x]  (phi) {$\phi$};
  \node[latent, right=of theta]  (tphi) {$\tilde \phi$};

  % Connect the nodes
  \edge {x,theta} {y} ; %
  \edge {phi}     {x} ; % 
  \edge{phi}      {theta}  % 
  \edge{tphi}      {theta}  % 
  
  \edge {tx,theta} {ty} ; %
  \edge {tphi}     {tx} ; % 

  % Plates
  \plate {yx} {(x)(y)} {$N$} ;
  \plate {txy} {(tx)(ty)} {$M$} ;
\end{tikzpicture}
  \caption{\footnotesize Our proposed probabilistic model for adaptation for covariate shift, observing a training set with $N$ labeled pairs and $M$ unlabeled inputs from a shifted test distribution.}
    
  \label{fig:shift_gm}
  \vspace{-12pt}
\end{wrapfigure}
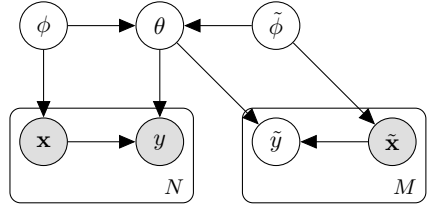
We now assume we have two input-generating distributions; $\phi$ specifies the input distribution for our labeled training set, and $\tilde \phi$ specifies the shifted distribution of unlabeled test inputs which we aim to adapt to.
Under the assumption of covariate shift, the same classifier $\theta$ is used to generate labels in both the train and test domains, leading to the model in Figure \ref{fig:shift_gm}.

We argue that our prior belief of low aleatoric uncertainty is reasonable for \emph{both} the distribution induced by $\phi$ and the one induced by $\tilde \phi$; even if there is some distribution shift from the training set, we can still often expect the test data we are shown to have unambiguous labels. 
We can then incorporate both $\phi$ and $\tilde \phi$ into our belief over $\theta$ with the factor 
\begin{align} \label{eq:ent-prior-da}
    p(\theta \vert \phi, \tilde \phi) &\propto \mu(\theta) \exp(-\alpha H_{\theta, \phi}(Y\vert \X))\exp(-\tilde \alpha H_{\theta, \tilde \phi}(Y\vert \X)),
\end{align}
with the two scalar hyperparameters $\alpha, \tilde \alpha$ controlling how to weight the entropies from each distribution with the likelihoods of the labeled training set. 
While we can extend this model to include more labeled or unlabeled input distributions, we focus on simply having one training distribution and one test distribution as it is the most relevant for our problem setting of adapting to a particular distribution shift at test time.

\subsection{Approximations for Tractable Inference}
Performing inference over $\theta$ in the above model can be challenging in our test-time adaptation setting for several reasons.
First of all, inference over $\theta$ would also require performing inference over the parameters of the generative models $\phi$ and $\tilde \phi$ to evaluate likelihoods in Equation \ref{eq:ent-prior-da}. 
This is difficult in practice for two reasons: First, the inputs $x$ might be high-dimensional and difficult to model, as in the case of images. Second, the amount of unlabeled data might be fairly small, and insufficient to accurately estimate the parameters $\tilde{\phi}$ if we employ a highly expressive generative model. As such, we would much prefer to avoid explict generative modeling and to only perform inference over the discriminative model parameters $\theta$ instead.

Another issue is that performing inference would require access to both the labeled training data and the unlabeled test data at the same time, while our test-time adaptation setting assumes that we can no longer look at the training set when it is time to adapt to the test distribution. We will discuss how to address these issues and describe our method, Bayesian Adaptation for Covariate Shift (BACS), which provides a practical instantiation of inference in this Bayesian model in a computationally tractable test-time adaptation setting.

\textbf{Plug-in approximation with empirical Bayes}.
We first propose to avoid explicit generative modeling by using with a plug-in empirical Bayes approach.
Empirical Bayes is a common approximation for hierarchical Bayesian models that simply uses a point estimate of $\phi$ rather than marginalizing over $\phi$'s (similarly for $\tilde \phi$), which would reduce the computation to estimating only a single generative model for each input distribution given by parameters $\phi^*$ and $\tilde \phi^*$.
To eliminate the need to train the parameters $\phi^*$ of a generative model of the inputs altogether,
we note that $p(\theta\vert \phi, \tilde \phi)$ only depends on $\phi$ and $\tilde \phi$ through the input distributions $p(x\vert \phi), p(\tilde x\vert \tilde \phi)$. 
We can then approximate Equation $\ref{eq:ent-prior-da}$ by plugging in the empirical distributions of $x$ and $\tilde x$ in place of $p(x\vert \phi^*), p(\tilde x\vert \tilde \phi^*)$, resulting in
\begin{align*}
    p(\theta\vert \hat \phi, \tilde \phi) \propto \mu(\theta) \exp\left(-\frac{\alpha}{n}\sum_{i=1}^n H(Y_i\vert \x_i, \theta) \right)\exp\left(-\frac{\tilde \alpha}{m}\sum_{i=1}^m H(Y_i\vert \tilde \x_i, \theta) \right).
\end{align*}
Now, given a labeled training set $\mathcal D$ and unlabeled test points $\mathcal U = \{\tilde \x_i\}_{i=1}^m$, the new posterior distribution over parameters now has log probabilities (up to an additive normalizing constant)
\begin{align}
    \log p(\theta \vert \mathcal D, \mathcal U) = \log \mu(\theta) + \sum_{i=1}^n \log p(y_i \vert \x_i, \theta) - \frac{\alpha}{n} \sum_{i=1}^n H(Y\vert \x_i, \theta) - \frac{\tilde \alpha}{m} \sum_{j=1}^m H(Y\vert \tilde \x_j, \theta).
\end{align}
For convenience, we simply set $\alpha= 0$, as additionally minimizing entropy on the labeled training set is unnecessary when we are already maximizing the likelihood of the given labels with highly expressive models.
Now, to infer $\theta$ given the observed datasets $\mathcal D$ and $\mathcal U$, the simplified log-density is 
\begin{align} \label{eq:ent-prior-simple}
    \log p(\theta \vert \mathcal D, \mathcal U) = \log \mu(\theta) + \sum_{i=1}^n \log p(y_i \vert \x_i, \theta) - \frac{\tilde \alpha}{m} \sum_{j=1}^m H(Y\vert \tilde \x_j, \theta).
\end{align}

Computing the maximum-a-posteriori (MAP) solution of this model corresponds to simply optimizing model parameters $\theta$ that on the supervised objective on the labeled training set in addition a minimum entropy regularizer on the unlabeled input. 
However, we should not necessarily expect the MAP solution to provide reasonable uncertainty estimates, as the learned model is being encouraged to make confident predictions on the test inputs, and so any single model will likely provide overconfident predictions.
Marginalizing the predictions over the posterior distribution over parameters is thus essential to recovering meaningful uncertainty estimates, as the different models, though each being individually confident, can still express uncertainty through their combined predictions when the models predict different labels.

\begin{algorithm}[t]
\caption{Bayesian Adaptation for Covariate Shift (BACS)}
\label{alg:bayes-ent-min}
\begin{algorithmic}
\STATE \textbf{Input}: Ensemble size $k$, Entropy weight $\tilde \alpha$, Training Data: $\x_{1:n}, y_{1:n}$, Test Data: $\tilde \x_{1:m}$
\STATE \textbf{Output}: Predictive distributions $p(y \vert \tilde \x_j)$ for each test input $\tilde \x_j$
\STATE \textbf{Training}: For each $i \in (1, \ldots, k)$ compute an approximate density  $q_i(\theta) \approx p(\theta \vert \x_{1:n}, y_{1:n})$
\STATE \textbf{Adaptation}: 
\FORALL{ensemble members $i \in (1, \ldots, k)$}
\STATE    Compute adapted parameters $\hat \theta_i = \argmax_{\theta} \frac{\tilde \alpha}{m}\sum_{j=1}^m -H(Y\vert \tilde \x_j, \theta) + \log q_i(\theta)$ \\
\ENDFOR
\STATE For each test input $\tilde \x_j$,  marginalize over ensemble  $p(y\vert \tilde \x_j) = \frac{1}{k} \sum_{i=1}^k p(y\vert \tilde \x_j, \hat \theta_i)$.
\end{algorithmic}
\end{algorithm}

\textbf{Approximate inference for test-time adaptation}. 
We now discuss how to perform inference in the above model in a way suitable for \textit{test-time adaptation}, where we adapt to the test data without any further access to the original training set.
To enable this, we propose to learn an approximate posterior 
\begin{align}
    q(\theta) \approx p(\theta \vert \mathcal D) = \log \mu(\theta) + \sum_{i=1}^n \log p(y_i\vert \x_i, \theta)
\end{align}
during training time, and then use this approximate training set posterior $q(\theta)$ in place of the training set when performing inference on the unlabeled test data.
This gives us an approximate posterior with log density
\begin{align}\label{eq:streaming-bayes}
    \log p(\theta \vert \mathcal D, \mathcal U) = \log q(\theta) - \frac{\tilde \alpha}{m} \sum_{j=1}^m H(Y\vert \tilde \x_j, \theta)
\end{align}
Unlike the full training set, the learned approximate posterior density $q(\theta)$, which can be as simple as a diagonal Gaussian distribution, can be much easier to store and optimize over for test-time adaptation, and the training time procedure would be identical to any approximate Bayesian method that learns a posterior density.
In principle, we can now instantiate test-time adaptation by running any approximate Bayesian inference algorithm, such as variational inference or MCMC, to sample $\theta$'s from the density in Equation \ref{eq:streaming-bayes}, and average predictions from these samples to compute the marginal probabilities for each desired test label.

\textbf{Practical instantiation with ensembles.}
As previously mentioned, marginalizing over different models that provide diverse labelings of the test set is crucial to providing uncertainty estimates after adaptation via entropy minimization.
We thus propose to use an ensembling approach \citep{Lakshminarayanan2016SimpleEnsembles} as a practical method to adapt to the test distribution while maintaining diverse labelings.
Deep ensembles simply train multiple models from different random initializations, each independently optimizing the target likelihoods, and averages together the models' predicted probabilities at test time. 
They are able to provide effective approximations to Bayesian marginalization due to their ability to aggregate models across highly distinct modes of the loss landscape \citep{Fort2019DeepPerspective, Wilson2020BayesianGeneralization}. 

Our method, BACS, summarized in Algorithm \ref{alg:bayes-ent-min}, trains an ensemble of $k$ different models on the training set, each with their approximate posterior $q_i(\theta)$ that captures the local loss landscape around each mode in the ensemble. 
Then at test time, we independently optimize each of the $k$ models by minimizing Equation \ref{eq:streaming-bayes} (using the corresponding $q_i(\theta)$ for each ensemble member), and then average the predictions across all adapted ensemble members.

%% file: related_work.tex
\section{Related Work}

\textbf{Entropy minimization}. Entropy minimization has been used as a self-supervised objective in many settings, including domain adaptation \citep{Saito2019Semi-supervisedEntropy, Carlucci2017AutoDIAL:Layers}, semisupervised learning \citep{Grandvalet2004Semi-supervisedMinimization, Berthelot2019MixMatch:Learning, Lee2013Pseudo-Label:Networks}, and few-shot learning \citep{Dhillon2015AClassification}.
\citet{Grandvalet2004Semi-supervisedMinimization} proposed a probabilistic model incorporating entropy minimization for semisupervised learning (without distribution shift),
but only use the probabilistic model to motivate entropy minimization as a regularizer for a MAP solution in order improve accuracy, which does not capture any epistemic uncertainty.
In contrast, we are concerned with test-time adaptation under distribution shift, which requires introducing a model of separate training-time and test-time input distributions, and with providing reliable epistemic uncertainty estimates, which we obtain via Bayesian marginalization.We also devise an approximate inference scheme to allow for efficient adaptation without access to the training data.

Test time entropy minimization (TENT) \citep{Wang2020Tent:Minimization} uses entropy minimization as the sole objective when adapting to the test data (though without an explicit Bayesian interpretation) and adapts without any further access to the training data, but only aims to improve accuracy and not uncertainty estimation. 
Similarly to \citet{Grandvalet2004Semi-supervisedMinimization}, TENT only learns a single model using entropy minimization, whereas we show that explicitly performing Bayesian inference and marginalizing over multiple models is crucial for effective uncertainty estimation. 
TENT also heuristically proposes to only adapt specific parameters in the networks at test time for stability reasons, while our usage of a learned posterior density to account for the training set allows us to adapt the whole network, improving performance in some settings and eliminating the need for the heuristic design decision.

\textbf{Domain Adaptation}. Unsupervised domain adaptation \citep{Ganin2015Domain-AdversarialNetworks, Wang2018DeepSurvey} tackles a similar problem of learning classifiers in the presence of distribution shift between our train and test distributions. Most unsupervised domain adaptation works assume access to both the labeled training data as well as unlabeled test data at the same time, while we restrict adaptation to occur without any further access to the training data. 
One recent line of work, known as source-free domain adaptation \citep{Liang2020DoAdaptation, Kundu2020UniversalAdaptation, Li2020ModelData} also restricts the adaptation procedure to not have access to the training data together with the unlabeled data from the test distribution.
In contrast to these algorithms, we are concerned with using adaptation to improve both uncertainty estimation as well as accuracy, and our algorithm is additionally amenable to an online setting, where prediction and adaptation occur simultaneously without needing to see the entirety of the test inputs.

\textbf{Uncertainty estimation under distribution shift (no adaptation)}. Various methods have been proposed for uncertainty estimation with distribution shift that do not incorporate test time adaptation. \citet{Ovadia2019CanShift} measured the calibration of various models under various distribution shift without any adaptation, finding that deep ensembles \citep{Lakshminarayanan2016SimpleEnsembles} and some other Bayesian methods that marginalize over multiple models perform well compared to methods that try to recalibrate predictions using only the source data. 
Beyond ensembles, other Bayesian techniques \citep{Dusenberry2020EfficientFactors, Maddox2019ALearning} have also demonstrated improved uncertainty estimation under distribution shift compared to standard models. 
Various other techniques have been found to improving calibration under distribution shift through significant changes to the training procedure, for example utilizing different loss functions \citep{Padhy2020RevisitingNetworks, Tomani2020TowardsCalibration}, extensive data-augmentations \citep{Hendrycks2019AugMix:Uncertainty}, or extra pre-training \citep{Xie2019Self-trainingClassification}. 

\textbf{Uncertainty estimation with both train and test distributions}. We now discuss prior work that considers uncertainty estimation assuming access to both train and test data simultaneously. Recent work studying calibration for domain adaptation algorithms \citep{Pampari2020UnsupervisedShift, Park2020CalibratedAdaptation, Wang2020TransferableAdaptation} found that predictions are poorly calibrated in the target domain even if the models were well-calibrated on the labeled source domain. 
These works all propose methods based on importance weighting between the target domain and the labeled source domain in order to recalibrate target domain predictions using only labels in the source domain. They are not directly applicable in our test-time adaptation setting, since they require estimates of density ratios between target and source distributions, which we cannot obtain without either a generative model of training inputs, or access to the training data during adaptation.
Our method also differs in how we approach uncertainty estimation. Instead of using extra labeled data and post-hoc recalibration techniques for classifiers, our method uses Bayesian inference to provide meaningful uncertainty estimates. 

For uncertainty estimation in regression problems, \citet{Chan2020UnlabelledShift} also propose to adapt Bayesian posteriors to unlabeled data by optimizing the predictive variance of Bayesian neural network at an input to serve as a binary classifier of whether the point is in-distribution or not. thus encouraging higher variance for out-of-distribution points. 
Their method is again not applicable in our test-time setting because they require access to both the train and test data at once.

\textbf{Uncertainty estimation with test time adaptation}.
\citet{Nado2020EvaluatingShift} evaluate various techniques for uncertainty estimation in conjunction with adapting batch-norm statistics to the shifted test domain, and again find that deep ensembles provide well-calibrated prediction in addition to improved accuracy.
While our method similarly utilizes ensembles and adapts batch norm statistics, and we show that additionally adapting via entropy minimization at test time further improves predictive accuracy without sacrificing calibration.

\textbf{Bayesian semi-supervised methods}: 
\citet{Seeger2000Input-dependentModels} proposed a probablistic model for incorporating unlabeled data in semi-supervised learning to motivate regularization for the classifier that depends on the input distribution in MAP inference. However, they do not tackle uncertainty estimation and their model does not account for any  distribution shift like ours does.
\citet{Gordon2020CombiningLearning} perform Bayesian semi-supervised learning combining both generative and discriminative models. In contrast, our method does not need to learn a generative model of the data, and explicitly tackles the problem of distribution shift instead of assuming the labeled and unlabeled data come from the same distribution.
Another line of work \citep{Ng2018BayesianProcesses, Ma2019ALearning, Walker2019GraphProcesses, Liu2020UncertaintyLearning} propose Bayesian methods for semisupervised learning specialized graph-structured data.

%% file: experiments.tex
\section{Experiments}
In our experiments, we aim to analyze how our test-time adaptation procedure in BACS performs when adapting to various types of distribution shift, in comparison to prior methods, in terms of \emph{both} the accuracy of the adapted model, and its ability to estimate uncertainty and avoid over-confident but incorrect predictions. We evaluate our method and prior techniques across a range of distribution shifts, including corrupted datasets, natural distribution shifts, and domain adaptation settings. 

\noindent \textbf{Architectures and implementation.} For our ImageNet \citep{Deng2009Imagenet:Database} experiments, we use the ResNet50v2 \citep{He2016IdentityNetworks} architecture, while for other datasets, we use ResNet26 \citep{He2016DeepRecognition}. 
For all methods utilizing ensembles, we use ensembles of 10 models, and report results averaged over the same 10 seeds for the non-ensembled methods.
While adapting our networks using the entropy loss, we also allow the batch normalization statistics to adapt to the target distribution.
To obtain approximate posteriors for each ensemble member, we use SWAG-D \citep{Maddox2019ALearning}, which estimates a Gaussian posterior with diagonal covariance from a trajectory of SGD and requires minimal changes to standard training procedures. During adaptation, we initialize each model from the corresponding posterior mean, corresponding to the solution obtained by Stochastic Weight Averaging \citep{Izmailov2018AveragingGeneralization}. For methods that optimize on the test distribution, we report results after one epoch of adaptation unless otherwise stated.

\noindent \textbf{Comparisons.} We compare our method against two state-of-the-art prior methods for test-time adaptation: TENT \citep{Wang2020Tent:Minimization}, which simply minimizes entropy on the test data with a single model, without the additional posterior term accounting for the training set that we use in BACS, and ensembles adapted using the batch norm statistics of the shifted test set, as discussed by \citet{Nado2020EvaluatingShift}. 
We also compare to deep ensembles \citep{Lakshminarayanan2016SimpleEnsembles} without any adaptation as a baseline for uncertainty estimation under distribution shift, as well as ensembles of models each adapted using TENT. 

\noindent \textbf{Metrics.} In addition to accuracy, we also evaluate uncertainty estimation using the negative log likelihood (NLL), Brier score \citep{Brier1950VerificationProbability} and expected calibration error (ECE) \citep{Naeini2015ObtainingBinning} metrics.
NLL and Brier score are both proper scoring rules \citep{Gneiting2007StrictlyEstimation}, and are minimized if and only if the predicted distribution is identical to the true distribution. 
ECE measures calibration by binning predictions according to the predicted confidence and averaging the absolute differences between the average confidence and empirical accuracy within each bin. 
\begin{table}[t]
    \centering
    \small
    \begin{tabular}{|c|cccc | c c  c c|}
        \hline
        \multicolumn{1}{|c}{} & \multicolumn{4}{|c}{CIFAR10-C} & \multicolumn{4}{|c|}{CIFAR100-C} \\
        % \hline
        Method & Acc & NLL & Brier & ECE & Acc & NLL & Brier & ECE\\
        \hline
        Vanilla   & 59.90 & 1.892  & 0.6216 & 0.2489 &            35.72 & 4.271 & 0.9797 & 0.3883 \\
        BN Adapt  & 82.36 & 0.8636 & 0.2909 & 0.1204 &            57.58 & 2.294 & 0.6401 & 0.2377 \\
        TENT (1 epoch) & 84.29 & 0.7862 & 0.2629 & 0.1119 &       62.46 & 2.047 & 0.5828 & 0.2280 \\
        TENT (5 epoch) & 85.16 & 0.8483 & 0.2603 & 0.1191 &       63.46 & 2.199 & 0.5987 & 0.2592 \\
        BACS (MAP) (1 epoch) & 84.82 & 0.7808 & 0.2585 & 0.1119 &     63.05 & 2.090 & 0.5908 & 0.2411 \\
        BACS (MAP) (5 epochs) & 85.20 & 0.8075 & 0.2575 & 0.1144 &    63.53 & 2.175 & 0.6009 & 0.2551 \\
        \hline
        Vanilla Ensemble & 61.72 & 1.535 & 0.5431 & 0.1684 &                       38.66 & 3.343 & 0.8439 & 0.2462 \\
        Ensemble BN Adapt & 85.99 & 0.4722 & 0.2043 & 0.03229 &                    64.22 & 1.464 & 0.4793 & \textbf{0.0515} \\ 
        Ensemble TENT (1 epoch) & 87.28 & 0.4351 & 0.1867 & \textbf{0.02868} &   67.83 & \textbf{1.318} & 0.4392 & 0.05909 \\
        BACS (ours) (1 epoch) & \textbf{87.77} & \textbf{0.4260} & \textbf{0.1809} & 0.02986 &       \textbf{68.33} & 1.324 & \textbf{0.4360} & 0.06519 \\
        \hline
    \end{tabular}
    
    \vspace{5pt} 
    \caption{\textbf{CIFAR-10/100 Corrupted} results at the highest level of corruption, averaged over all corruption types. With one epoch of adaptation, BACS consistently outperforms all baselines in terms of accuracy, NLL and Brier score, and can further improve with more training. In terms of ECE, all ensembled methods with adaptation performs similarly to well, substantially outperforming non-adaptive or non-ensembled baselines.}
    \vspace{-20pt}
    \label{tab:cifar10/100}
\end{table}

\begin{figure}
    \centering
    \includegraphics[width=1.0\linewidth]{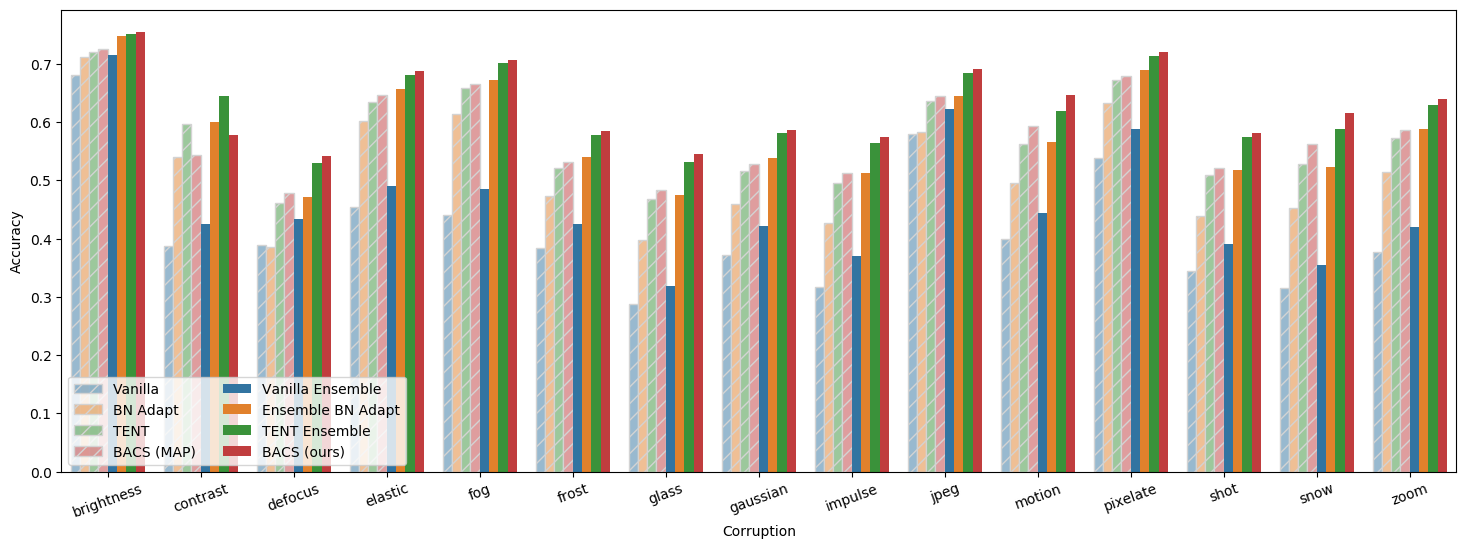} \\
    \vspace{-10pt}
    \includegraphics[width=1.0\linewidth]{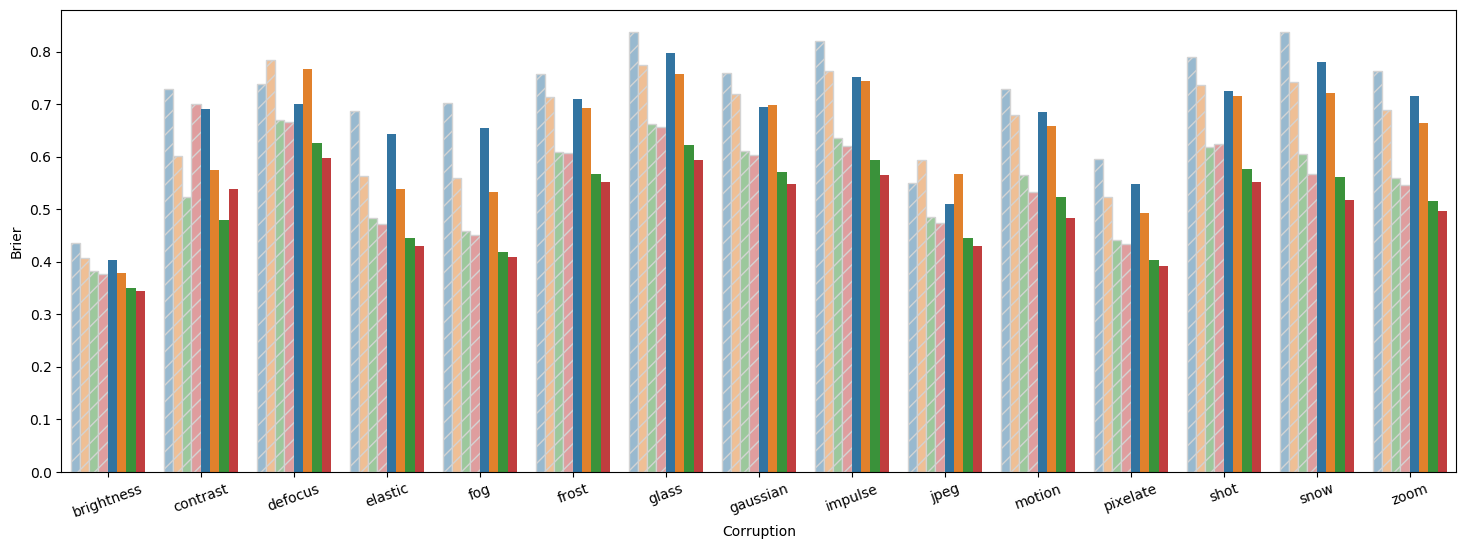}
    \vspace{-15pt}
    \caption{\textbf{ImageNet-C Results by Corruption}. For each corruption type, we show the results for each method averaged over all levels of corruption. BACS improves over baselines for every corruption type except for contrast, where BACS has close to 0 accuracy at the most severe level of corruption.}
    \label{fig:imagenetc-barplots-main}
    \vspace{-20pt}
\end{figure}

\textbf{Corrupted images.} We first evaluate our method on CIFAR-10-C, CIFAR-100-C, and ImageNet-C \citep{Hendrycks2019BenchmarkingPerturbations}, where distribution-shifted datasets are generated by applying different image corruptions at different intensities to the test sets of CIFAR10, CIFAR100 \citep{Krizhevsky2012LearningImages}, and ImageNet \citep{Deng2009Imagenet:Database}
respectively. 
In Table \ref{tab:cifar10/100}, we show comparisons at the most severe level of corruption for CIFAR10-C and CIFAR100-C.
With just a single epoch of adaptation at test time, both our method BACS and TENT ensembles substantially outperform other baselines in accuracy, NLL, and Brier score, with BACS improving slightly over TENT ensembles, showing that combining test-time entropy minimization with Bayesian marginalization can lead to strong improvements over either alone. 
We also note that simply adapting for more epochs with entropy minimization (as seen in TENT (5 epochs)), can further improve accuracy of a single model, but can actually lead to \textit{worse} uncertainty estimates as measured by NLL or ECE as predictions become excessively overconfident.
In contrast, combining ensembling with entropy minimization consistently improves accuracy much more than simply adapting a single model for more epochs, while also substantially improving uncertainty estimates.

\begin{wraptable}{r}{0.60\linewidth}
    \vspace{-15pt}
    \small
    \centering
    \begin{tabular}{|c|cccc| }
        \hline
        Method & Acc & NLL & Brier & ECE\\
        \hline
        Vanilla & 41.79 & 3.127 & 0.7152 & 0.06439\\ 
        BN Adapt  & 51.54 & 2.676 & 0.6564 & 0.1709\\ 
        TENT & 57.06 & 2.035 & 0.5536 & \textbf{0.0342}  \\ 
        BACS (MAP)& 58.06 & 2.036 & 0.5528 & 0.04270 \\
        \hline
        
        Vanilla ensemble & 46.03  & 2.788 & 0.6670 & 0.07256 \\ 
        
        Ensemble BN Adapt & 58.30 & 2.428 & 0.6333 & 0.2594 \\
        Ensemble TENT & 62.40 & 1.788 & 0.5131 & 0.1010\\
        BACS (ours) & \textbf{63.04} & \textbf{1.726} & \textbf{0.4962} & 0.08308\\
        \hline
    \end{tabular}
    \vspace{-5pt}
    \caption{\textbf{ImageNet-C} results averaged over all corruption types and levels. 
    BACS outperforms all baselines in accuracy, NLL, and Brier scores.}
    \vspace{-5pt}
    \label{tab:imagenet-c}
\end{wraptable} 
In Table \ref{tab:imagenet-c}, we show comparisons on ImageNet-C averaged over all corruption levels and corruption types. 
BACS is able to outperform all prior methods in terms of accuracy, NLL, and Brier score. We note that ensembling and batch norm adaptation actually \textit{worsen calibration}, as measured by ECE, compared to the single model or non-adapted baselines respectively, despite each technique providing substantial improvements on all other metrics.
We see that an ablation of our method that only uses a single model, BACS (MAP), outperforms the other non-ensemble methods in accuracy, NLL, and Brier score.

We also show results for accuracy and Brier score at each level of corruption in Figure \ref{fig:imagenetc-barplots-main}. BACS (ours) consistently outperforms baselines at each level of corruption (with one exception being accuracy at the highest level of corruption, where a single corruption with much lower accuracy drags down the mean to be slightly below to that of Ensemble TENT). 

\textbf{ImageNet-R.} In Table \ref{tab:imagenet-R}, we further evaluate robustness using ImageNet-R \citep{Hendrycks2021TheGeneralization}, which consists of images that are abstract renditions of 200 of the ImageNet classes. Similar to our Imagenet-C results, we find that BACS performs the best across accuracy, NLL, and Brier score, while being slightly outperformed in ECE by vanilla ensembles.

\begin{wraptable}{r}{0.60\linewidth}
    % \vspace{-12pt}
    \small
    \centering
    \begin{tabular}{|c|cccc| }
        \hline
        Method & Acc & NLL & Brier & ECE\\
        \hline
        Vanilla & 36.40 & 3.288 & 0.7602 & 0.05667 \\ 
        BN Adapt  & 38.05 & 3.236 & 0.7646 & 0.09744  \\
        TENT & 41.13 & 2.967 & 0.7167 & 0.02666 \\
        BACS (MAP)& 43.55 & 2.909 & 0.7183 & 0.1074 \\
        \hline
        Vanilla ensemble & 40.38 & 3.011 & 0.7180 & \textbf{0.02339} \\ 
        Ensemble BN Adapt & 42.82 & 3.019 & 0.7408 & 0.1696 \\
        Ensemble TENT & 45.75 & 2.726 & 0.6815 & 0.1025 \\
        BACS (ours) & \textbf{47.31 }& \textbf{2.565} & \textbf{0.6625} & 0.04270 \\
        \hline
    \end{tabular}
    \vspace{-5pt}
    \caption{\textbf{ImageNet-R} results. 
    BACS outperforms all baselines in accuracy, NLL, and Brier scores.}
    \vspace{-10pt}
    \label{tab:imagenet-R}
\end{wraptable} 

\textbf{Impact of posterior term.} In this section, we discuss the effects of using the training set posterior density in our objective (Equation \ref{eq:streaming-bayes}) when adapting at test time. 
Intuitively, the training posterior density ensures that our adapted classifiers, while minimizing entropy on the target distribution, remain constrained to still perform well on the training set and stay near the initial solution found during training. 
We empirically find that the posterior term can be important for preventing entropy minimization from finding degenerate solutions: adapting the whole network on several ImageNet-C corruptions without the posterior term can lead to poor models that achieve close to 0 accuracy on many corruptions. 

While TENT, which adapts via entropy minimization without any term accounting for the training data, uses an ad-hoc solution that restricts adaptation to only the learnable scale and shift parameters in the batch norm layers, our use of the training set posterior density is motivated directly from our proposed probabilistic model and does not require any heuristic choices over which parameters should or should not be adapted.
In our ImageNet experiments (Tables 2 and 3), we see that an ablation of our method without ensembles, corresponding to the MAP solution in our proposed model, outperforms TENT. The difference between our ablation BACS (MAP) and TENT is precisely that BACS (MAP) adapts the whole network, but with an additional regularization term, while TENT adapts only a small part of the network without any regularizer.

However, we find the approximate posterior density can also be limiting in certain situations, which we can observe in experiments transferring from SVHN to MNIST (Appendix \ref{sec:da}).
Here, there is a large discrepancy between the training and test domains, and we find that adaptation with the posterior is unable to adjust the parameters enough and underperforms compared to an ablation of our method that simply removes the posterior term while still adapting the whole network.

%% file: conclusion.tex
\section{Discussion} 
We presented Bayesian Adaptation for Covariate Shift (BACS), a Bayesian approach for utilizing test-time adaptation to obtain both improved accuracy and well-calibrated uncertainty estimates when faced with distribution shift.
We have shown that adapting via entropy minimization without Bayesian marginalization can lead to overconfident uncertainty estimates, while our principled usage of an approximate training posterior during adaptation can outperform previous heuristic methods.
These observation support our hypothesis that framing entropy minimization within a well-defined Bayesian model can lead to significantly more effective test-time adaptation techniques.
Our method is straightforward to implement, requires minimal changes to standard training procedures, and improves both accuracy and uncertainty estimation for a variety of distribution shifts in classification. 

\noindent \textbf{Limitations and future work.} 
One limitation of our approach is that it requires effective techniques for estimating the parameter posterior from the training set. While the study of such methods, and Bayesian neural networks more broadly, is an active area of research, it remains a significant practical challenge. It is likely that the simple Gaussian posteriors we employ provide a poor estimation of the true posterior and can overly constrain the network during adaptation. Therefore, a relevant direction for future work is to integrate BACS with more sophisticated Bayesian neural network methods. 

Another promising direction for future work is to explore other objectives that have had success in semi-supervised learning settings, such as consistency based losses \citep{Sajjadi2016RegularizationLearning, Miyato2017VirtualLearning, Xie2019UnsupervisedTraining, Sohn2020FixMatch:Confidence} or information maximization \citep{Gomes2010DiscriminativeMaximization}, which can be straightforwardly incorporated into our method as suitable priors on the relationship between the data distribution and classifier. 
More broadly, we hope that our work will spur further research into test-time adaptation techniques based on well-defined Bayesian models that describe how unlabeled test data should inform our posterior estimates of the model parameters, even in the presence of distributional shift.

%% file: appendix.tex
\section{Experimental Details}
\subsection{Training Details}
\textbf{Model Training}: Following TENT \citep{Wang2020Tent:Minimization}, we use ResNet26 models \citep{He2016DeepRecognition} for all experiments besides the ImageNet experiments. For ImageNet, we use ResNet50v2 \citep{He2016IdentityNetworks}. All methods utilizing ensembles use 10 different models in the ensemble.

For all ResNet26 experiments, we train all models and obtain posteriors using SWAG-D \citep{Maddox2019ALearning} and use the same hyperparameters as provided by the authors for training a WideResNet28x10. 
For methods that do not include the posterior density term, we simply use the mean of the learned SWAG-D posterior, which simply corresponds to the solution obtained by Stochastic Weight Averaging (SWA) \citep{Izmailov2018AveragingGeneralization}.
We train for 300 epochs on each training dataset, starting to collect iterates for SWA/SWAG starting from epoch 160.
We use SGD with momentum for all experiments, with a learning rate schedule given by 0.1 for the first 150 epochs, decaying linearly down to 0.01 until epoch 270, then remaining at 0.01 for the remaining 30 epochs.

For ResNet50v2 experiments on ImageNet, we first train for 90 epochs using the example training script at \url{https://github.com/deepmind/dm-haiku/tree/master/examples/imagenet}. 
We then train for an additional 10 epochs with a constant learning rate of 0.0001 to collect iterates for SWAG, collecting 4 iterates per epoch.

\textbf{Adaptation Details}: For test time adaptation on the small scale experiments with ResNet26, we use the same hyperparameters described in TENT \citep{Wang2020Tent:Minimization}. 
using learning rate 0.001 and batch size 128, using SGD with momentum.
For the ImageNet experiments with ResNet50v2, we tune the learning rate for accuracy using the validation corruptions in ImageNet-C, and keep the batch size fixed at 64. For TENT, we found a learning rate 0.001 to perform the best, while for our method BACS, we found a smaller learning rate of 0.0001 to perform the best.

During adaptation with BACS, according to equation \ref{eq:streaming-bayes}, we would optimize each model with the objective
\begin{align}
    \max_{\theta} \log q_i(\theta) - \frac{\tilde \alpha}{m}\sum_{j=1}^m -H(Y\vert \tilde \x_j, \theta).
\end{align}
where $q_i(\theta)$ is a diagonal Gaussian posterior density obtained by SWAG-D, and $\tilde \alpha$ is a hyperparameter controlling how much to weight the entropy minimization objective against the posterior term.
We initialize from the mean of the SWAG posterior before adaptation.
In practice, we also rewrite the objective as 
\begin{align}
    \max_{\theta} \beta \log q_i(\theta) - \frac{1}{m}\sum_{j=1}^m -H(Y\vert \tilde \x_j, \theta),
\end{align}
where $\beta = \frac{1}{\tilde \alpha}$, and use minibatches of test inputs for the entropy minimization term.
For all small-experiments with ResNet26, we use $\beta = 0.0001$.
For the ImageNet experiments, we jointly tuned $\beta$ with the learning rate using the extra validation corruptions and selected $\beta = 0.0003$.

\textbf{Compute Resources}: 
The initial training for each ResNet50 model on ImageNet used a Google Cloud TPU v3 instance with 8 cores, each taking approximately 20 hours to train. In total, as we needed to train 10 models for the ensemble, this utilized 200 hours of compute with TPU v3 instances with 8 cores each.

All other model training and all the adaptation were run using local GPU servers with Nvidia Titan RTX GPUs. 
We estimate model training time to be around 200 hours in total for all ResNet26 models (10 models for each of CIFAR10, CIFAR100, and SVHN and an estimated 6 hour training time with on one GPU). 
For adaptation and evaluation, we estimate the total time needed for the ImageNet-Corrupted experiments, as these would take up the bulk of the time spent on adaptation. For each of the 15 corruption types and 5 corruption levels, we estimate the total time it takes to adapt each model using entropy minimization for one epoch, evaluate the model without any adaptation, evaluate the model with adapted batchnorm statistics, and evaluate each model after entropy minimization to be 10 minutes.
With 10 models for each corruption, this totals 125 hours of GPU time to compile the ImageNet Corrupted results, which take up the bulk of the time needed for adaptation and evaluation in our experiments.

\subsection{Metrics}
We compute Brier score for a probability vector $p(y\vert x_n)$ and true label $y_n$ as
\begin{align}
    \sum_{y\in \mathcal Y}(p(y\vert x_n) - \delta(y - y_n))^2. 
\end{align}

To compute expected calibration error (ECE), we use 20 bins. 
We order all predictions by confidence and aggregate them into 20 bins, each with the same number of data points, instead of using fixed windows for the buckets. 
Given the bins $B_i$, we then compute ECE as
\begin{align}
    \text{ECE} = \frac{1}{20} \sum_{i=1}^{20} \vert \text{acc}(B_i) - \text{conf}(B_i)\vert,
\end{align}
where acc and conf compute the average accuracies and confidences within the predictions in each bucket.

\subsection{Dataset details}
Models were trained on CIFAR10/100 \citep{Krizhevsky2012LearningImages}, ImageNet \citep{Deng2009Imagenet:Database}, CIFAR10/100-Corrupted and Imagenet-Corrupted \citep{Hendrycks2019BenchmarkingPerturbations}, STL10 \citep{Coates2011AnLearning}, SVHN \citep{Netzer2011Learning}, MNIST \citep{LeCun2010MNISTDatabase}.

For the corrupted datasets, we report results averaged over all 15 standard corruption types in \citep{Hendrycks2019BenchmarkingPerturbations}, using the 4 extra corruption types for validation.

CIFAR10-Corrupted, STL10, SVHN, and MNIST were downloaded using Tensorflow Datasets. The Imagenet training data was directly downloaded from \url{www.image-net.org}), while CIFAR-100-Corrupted and Imagenet-Corrupted were downloaded from the author-released images at \url{zenodo.org/record/3555552} and \url{https://zenodo.org/record/2235448} respectively. 

For Imagenet-C, we initially ran experiments and performed hyperparameter tuning using the datasets generated through Tensorflow Datasets, which recomputed the the corrupted images instead of simply downloading the images released by the original authors. 
However, performance on the TFDS version of ImageNet-C (see appendix \ref{sec:expanded-imagenet-results}) is not directly comparable to that using the official released dataset, with overall results being substantially stronger with the TFDS version. 
This discrepancy is also noted in \citet{Ford2019AdversarialNoise}, who postulated the performance differences are due to the extra JPEG compression used for the officially released dataset making the tasks more difficult.

We were unable to find licensing information for CIFAR10, CIFAR100 or STL10. 
Imagenet is released under a custom license stipulating non-commercial research and educational use only (see \url{www.image-net.org/download}).
SVHN is released with a note stating it should be used for non-commercial uses only. 
MNIST is released under the CC BY-SA 3.0 license.
CIFAR10/100-Corrupted and Imagenet-Corrupted are released under the CC BY 4.0 License.

\section{Additional Experimental Results}
\subsection{Online Evaluation}

\begin{table}[t!]
    \centering
    \small
    \begin{tabular}{|c|c|c|c|c | c| c | c |c|}
        \hline
        \multicolumn{1}{|c}{} & \multicolumn{4}{|c}{CIFAR10-C} & \multicolumn{4}{|c|}{CIFAR100-C} \\
        \hline
        Method & Acc & NLL & Brier & ECE & Acc & NLL & Brier & ECE\\
        \hline
        TENT  & 84.52 & 0.7248 & 0.2289 & 0.09914          & 64.37 & 3.265 & 0.6111 & 0.2759\\
        BACS (ours) & 87.43 & 0.4242 & 0.1845 & 0.02882     & 67.28 & 1.323 & 0.4435 & 0.0500\\
        \hline
    \end{tabular}
    
    \vspace{5pt} 
    \caption{\textbf{CIFAR-10/100 Corrupted Online} results at the highest level of corruption, averaged over all corruption types at severity level 5 (the most severe level). While online adaptation performs slightly worse than offline adaptation, our method BACS still provides substantial improvements over TENT, ensembles with BN adaptation and other baselines.}
    \label{tab:cifar10/100-online}
\end{table} 
For the results in the main paper, all methods that adapted to the test distribution were evaluated in an offline setting, where each method could access the full test dataset before making any predictions. 
For BACS, this involves first making one pass through the test data to update the network for one epoch of optimization (or multiple epochs when specified), then making one more pass through the dataset to make predictions with the fully adapted network.

As there might be scenarios where we do not have access to all the test data from a particular distribution at once, we also include experiments on the corrupted datasets evaluating methods in an online setting, where we adapt the network and make predictions in a single pass through the dataset. 
The online procedure is more computationally efficient (requiring one fewer forward pass per batch) and does not require the model to wait for all the data to arrive before making predictions.
Note that all adaptive methods we evaluate still require access to batches of test data to make updates, and we utilize the same adaptation hyperparameters as with the offline experiments.
For online experiments, we fix the ordering on the test dataset for all methods.

\begin{wraptable}{r}{0.63\linewidth}
    \vspace{-10pt}
    \small
    \centering 
    \begin{tabular}{|c|c|c|c|c| }
        \hline
        Method & Acc & NLL & Brier & ECE\\
        \hline
        TENT & 54.12  & 2.330 & 0.6005 & 0.09544  \\ 
        BACS (MAP)& 56.13 & 2.171 & 0.5771 & 0.07055  \\ 
        \hline
        TENT Ensemble & 60.22 & 2.093 & 0.5702 & 0.1828 \\
        BACS (ours) & 61.81 & 1.911 & 0.5369 & 0.1503 \\
        \hline
    \end{tabular}
    \vspace{-5pt}
    \caption{\textbf{ImageNet-C Online} results averaged over all corruption types and levels. 
    Again, online adaptation performs worse compared to offline adaptation, but online BACS still outperforms one TENT (as well as other baseliens) in accuracy, NLL, and Brier score.}
    \vspace{-25pt}
    \label{tab:imagenet-c-online}
\end{wraptable}
We show results in the online setting for TENT and BACS in Tables \ref{tab:cifar10/100-online} and \ref{tab:imagenet-c-online}. Compared to the offline results in Tables \ref{tab:cifar10/100} and \ref{tab:imagenet-c}, online BACS generally performs worse than offline BACS, but still outperforms all baselines in accuracy, NLL, and Brier score.

\subsection{In Distribution Results}
At test time, it is also possible that the distribution we encounter is actually the same as training, though we would not necessarily know a priori. 
We thus also evaluate the performance of different adaptive methods when there is no distribution shift at test time in Table \ref{tab:cifar10/100-indist}. 
We see that BACS does slightly underperform relative to ensembles without adaptation and ensembles with batch-norm adaptation.
\begin{table}[t]
    \centering
    \small
    \begin{tabular}{|c|cccc | c c  c c|}
        \hline
        \multicolumn{1}{|c}{} & \multicolumn{4}{|c}{CIFAR10} & \multicolumn{4}{|c|}{CIFAR100} \\
        \hline
        Method & Acc & NLL & Brier & ECE & Acc & NLL & Brier & ECE\\
        \hline
        Vanilla & 95.50 & 0.1715 & 0.07252 & 0.02549
    & 77.88 & 1.023 & 0.3389 & 0.1198
\\
        BN Adapt  & 95.49 & 0.1767 & 0.07303 & 0.02588
         & 77.88 & 1.071 & 0.3437  & 0.1266
\\
        TENT & 95.48 & 0.1788 & 0.07662 & 0.03180
          & 77.86 & 1.083 & 0.3458 & 0.1322
\\
        BACS (MAP) & 95.40 & 0.1827 & 0.07734 & 0.02884 & 77.81 & 1.1388 & 0.3494 & 0.1412
        \\
        \hline
        
        Vanilla ensemble & 96.07 & \textbf{0.122} & \textbf{0.05835} & 0.009644
     & 80.34 & \textbf{0.7182} & \textbf{0.2711}& \textbf{0.04254} \\
        Ensemble BN Adapt & \textbf{96.09} & 0.1244 & 0.05834 & \textbf{0.009433}
       & 80.46& 0.7312 & 0.2720 & 0.04438 \\
       TENT Ensemble & 96.08 & 0.1240 & 0.05869 & 0.01065 &\textbf{80.68}& 0.7341 & 0.2719 & 0.05352\\ 
        BACS (ours)  & 95.98 & 0.1340 & 0.06154 & 0.01091
     & 80.32 & 0.7428 & 0.2754 & 0.05051
\\
        \hline
    \end{tabular}
    
    \vspace{5pt} 
    \caption{\textbf{CIFAR-10/100 In-distribution results.} We evaluate all methods on the standard (uncorrupted) test sets.}
    \label{tab:cifar10/100-indist}
\end{table} 

\subsection{Expanded ImageNet-C Results} \label{sec:expanded-imagenet-results}

\begin{wraptable}{r}{0.50\linewidth}
    \vspace{-10pt}
    \small
    \centering
    \begin{tabular}{|c|c| }
        \hline
        Method & mCE\\
        \hline
        Vanilla & 73.42 \\
        BN Adapt  & 61.24\\
        TENT & 54.37 \\ 
        
        BACS (MAP)& 53.13  \\ 
        \hline
        
        Vanilla Ensemble & 67.98 \\ 
        Ensemble BN Adapt & 52.70 \\
        TENT Ensemble & 47.50 \\
        BACS (ours) & \textbf{46.78} \\
        \hline
    \end{tabular}
    \vspace{-5pt}
    \caption{\textbf{ImageNet-C mCE} results. BACS (ours) outperforms all other methods, while our ablation without ensembles BACS (MAP) outperforms all non-ensembled methods.}
    \vspace{-15pt}
    \label{tab:imagenet-c-mce}
\end{wraptable}

In addition to measuring the average accuracy across the ImageNet-C corruptions, we also report results using the \textit{mean corruption error} (mCE) \citep{Hendrycks2019BenchmarkingPerturbations}, which normalizes the per-corruption error rates using the performance of an AlexNet model as a baseline before averaging across corruption types. 
We include these results in Table \ref{tab:imagenet-c-mce}. 

\begin{figure}
    \centering
    \includegraphics[width=\linewidth]{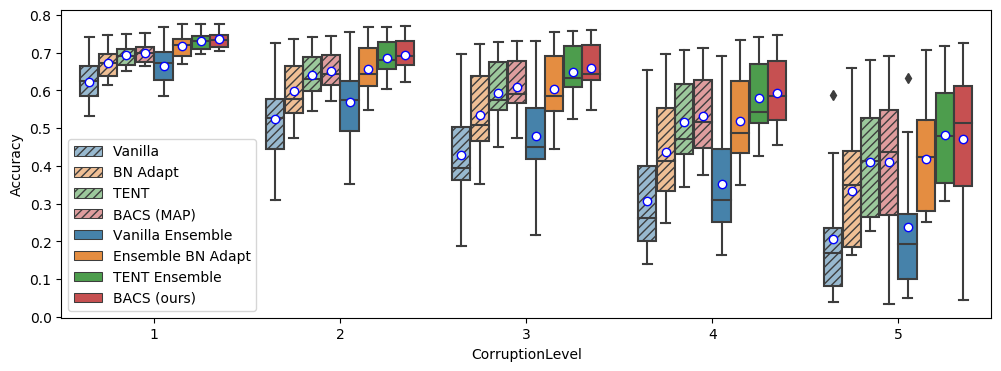} \\
    
    \includegraphics[width=\linewidth]{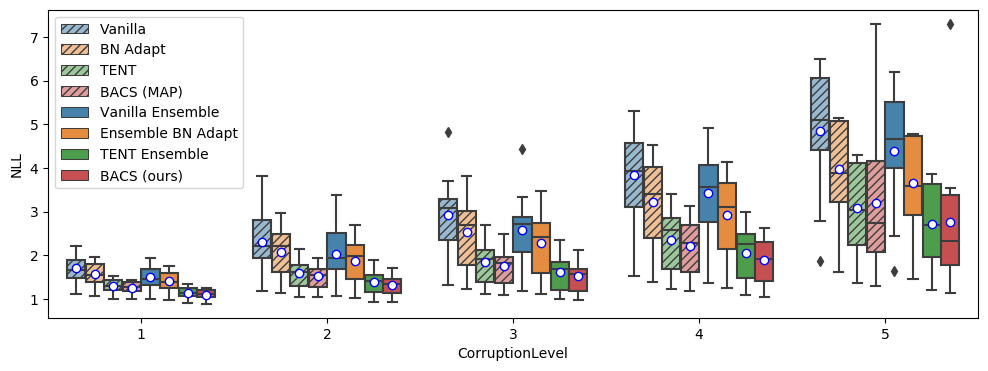}\\
    
    \includegraphics[width=\linewidth]{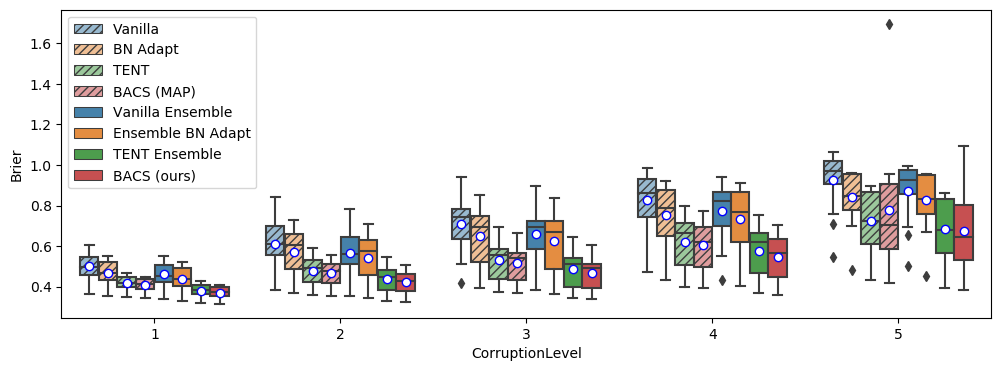}
    
    \includegraphics[width=\linewidth]{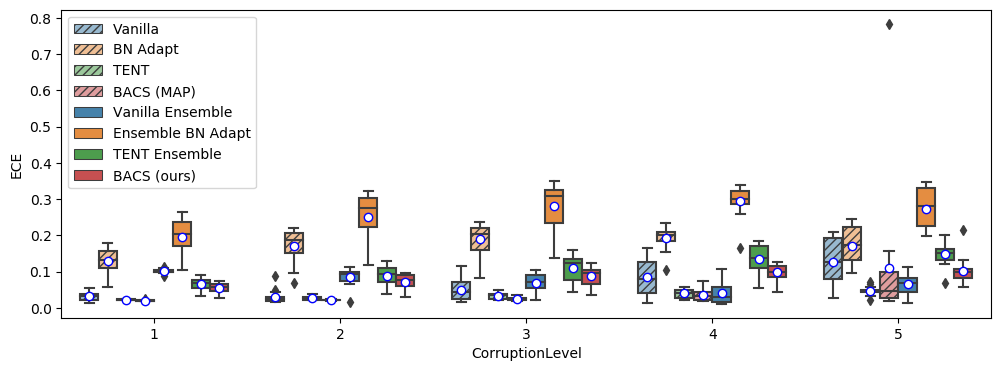}
    \caption{\textbf{ImageNet Corrupted Results}. For each corruption level, we use boxplots the spread of results over the different individual corruption types. Boxes are drawn at the 25th and 75th percentiles, with the median being drawn as a line in the middle of the box and the mean being shown with white dots. The ends of the whiskers show the min and max across corruption types at the level (with black diamonds for outliers).}
    \label{fig:imagenetc-boxplots}
\end{figure}

We include expanded experimental results for ImageNet-Corrupted in Figure \ref{fig:imagenetc-boxplots}. For each metric, we now use box plots to show the variability of results over different corruption types, separating out results at each level of corruption.
Across all corruption levels, BACS consistently performs the best in accuracy, NLL and Brier score (with the exception being the mean accuracy at the highest level of corruption, where a single corruption where BACS has very low accuracy drags down the mean, though median performance is still higher than all baselines).

\begin{wraptable}{r}{0.60\linewidth}
    \vspace{-10pt}
    \small
    \centering
    \begin{tabular}{|c|cccc| }
        \hline
        Method & Acc & NLL & Brier & ECE\\
        \hline
        Vanilla & 45.64 & 2.867 & 0.6750 & \textbf{0.0614}  \\ 
        BN Adapt  & 55.97 & 2.363 & 0.6050 & 0.1606 \\
        TENT & 60.82  & 1.803 & 0.522 & 0.0313  \\ 
        
        BACS (MAP)& 61.96 & 1.712 & 0.5022 & 0.0294  \\ 
        
        \hline
        Vanilla Ensemble & 50.48  & 2.519 & 0.6211 & 0.07878 \\
        Ensemble BN Adapt & 62.18 & 2.137 & 0.5802 & 0.2438 \\
        TENT Ensemble & 65.83 & 1.586 & 0.4726 & 0.0956 \\ 
        BACS (ours) & \textbf{66.64} & \textbf{1.492} & \textbf{0.4548} & 0.0735 \\
        \hline
    \end{tabular}
    \vspace{-5pt}
    \caption{\textbf{ImageNet-C (TFDS)} results averaged over all corruption types and levels. 
    BACS again substantially outperforms all baselines in accuracy, NLL, and Brier score.}
    \vspace{-15pt}
    \label{tab:imagenet-c-tfds}
\end{wraptable} 

\begin{figure}
    \centering
    \includegraphics[width=\linewidth]{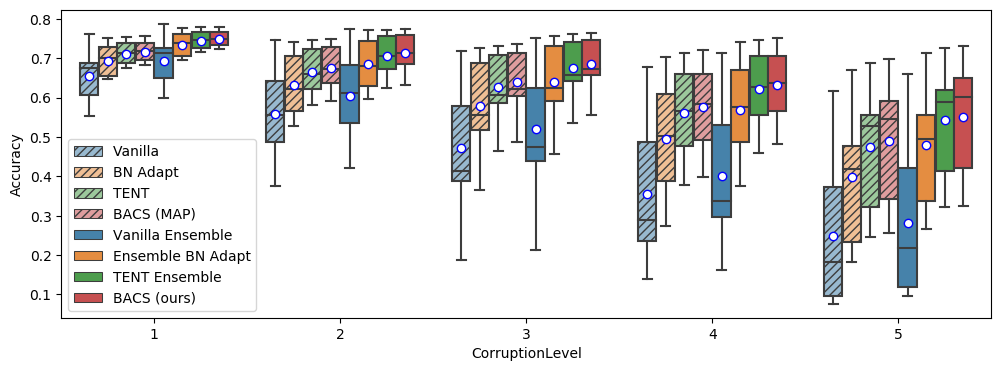} \\
    
    \includegraphics[width=\linewidth]{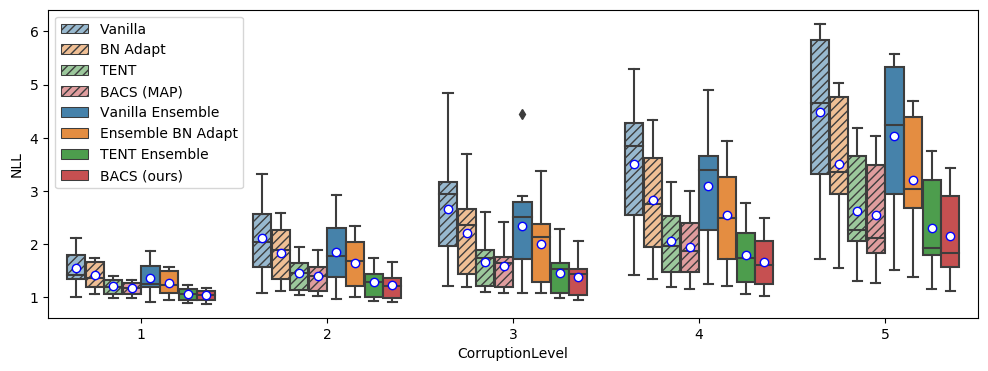}\\
    
    \includegraphics[width=\linewidth]{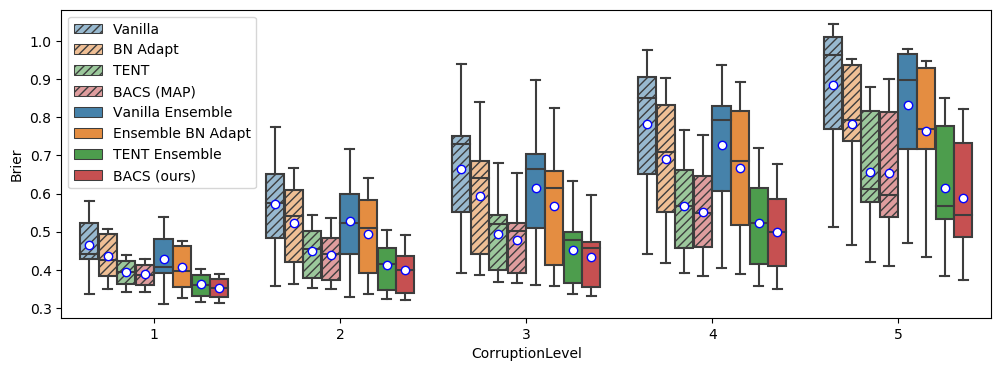}
    
    \includegraphics[width=\linewidth]{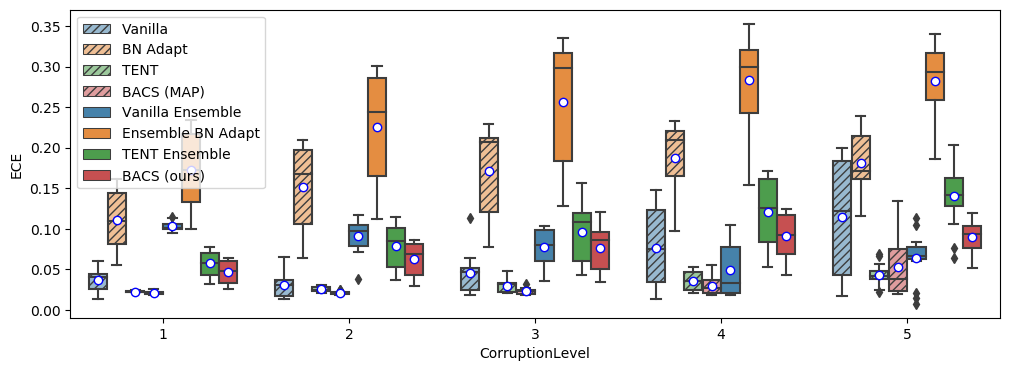}
    \caption{\textbf{ImageNet Corrupted (TFDS) Results}. For each corruption level, we use boxplots the spread of results over the different individual corruption types. Boxes are drawn at the 25th and 75th percentiles, with the median being drawn as a line in the middle of the box and the mean being shown with white dots. The ends of the whiskers show the min and max across corruption types at the level (with black diamonds for outliers). At each corruption level, BACS (ours) outperforms all baselines in accuracy, NLL, and Brier score.}
    \label{fig:imagenetc-tfds-boxplots}
\end{figure}

We also include experimental results using the TFDS version of ImageNet-C in Table \ref{tab:imagenet-c-tfds}, which we note has substantially better results overall than the officially released dataset. 
We also included boxplots for the Tensorflow Datasets version of ImageNet-Corrupted in Figure \ref{fig:imagenetc-tfds-boxplots}, where we see BACS performs the best in accuracy, NLL, and Brier scores at each level of corruption.

\subsection{Domain Adaptation Results} \label{sec:da}
We further evaluate BACS in additional small-scale experiments commonly evaluated for domain adaptation. 
We evaluate transferring from CIFAR10 to STL10 (using only the 9 overlapping classes), as well as a commonly used digit recognition task transferring from SVHN to MNIST.

\begin{wraptable}{r}{0.62\linewidth}
    % \vspace{-10pt}
    \centering
    \small
    \begin{tabular}{|c|cccc|}
        \hline
        Method & Acc & NLL & Brier & ECE\\
        \hline
        Vanilla & 82.38 & 0.7825 & 0.2886 & 0.1185 \\
        BN adapt & 83.72 & 0.7926 & 0.2709 & 0.1147\\
        TENT & 84.05 & 0.8831 & 0.2753 & 0.1216 \\
        
        \hline
        
        Vanilla Ensemble & 84.03 & 0.5360 & 0.2333 & 0.06561\\
        Ensemble BN Adapt & 85.40 & 0.5301 & 0.2195 & \textbf{0.05949}\\
        TENT Ensemble & 85.10 & 0.5337 & 0.2270 & 0.06844 \\ 
        BACS (ours)  & \textbf{85.47} & \textbf{0.5284} & \textbf{0.2184} & 0.06114\\
        \hline
    \end{tabular}
    \caption{\textbf{CIFAR10 to STL10}: Source model trained on CIFAR10 and evaluated on STL10 test set the (with nonoverlapping classes removed during both training and test). Here, the ensembled approaches perform the best overall and perform similarly to one another in uncertainty estimation.}
    \vspace{-12pt}
    \label{tab:cifar_stl}
\end{wraptable}
\textbf{CIFAR10 to STL10.} We further evaluate all methods on a more natural distribution shift by considering evaluating a model trained CIFAR-10 and evaluated on STL-10 \citep{Coates2011AnLearning}. We only use the 9 overlapping classes in each dataset. 
In this setting, BACS and BN ensembles outperform the other methods.

While TENT is able to improve accuracy slightly over the non-ensembled baselines, we again see that uncertainty estimates degrade as entropy is minimized (as seen by the increases in NLL, Brier, and ECE scores). In contrast, BACS still remains well-calibrated, again emphasizing the importance of Bayesian marginalization for reliable uncertainty estimation when adapting models via entropy minimization.

\textbf{SVHN to MNIST transfer}.
We also evaluate our method on a digit classification task, transferring a model trained on SVHN \citep{Netzer2011Learning} to MNIST \citep{LeCun2010MNISTDatabase}, which is commonly studied as a the domain adaptation setting \citep{Ganin2015Domain-AdversarialNetworks, Liang2020DoAdaptation}.
We find that BACS is able to significantly outperform the models with no adaptation as well as ensembles with batch norm adaptation in accuracy, NLL, and Brier score.

\begin{table} 
    \vspace{-10pt}
    \centering
    \small
    \begin{tabular}{|c|cccc|}
        \hline
        Method & Acc & NLL & Brier & ECE\\
        \hline
        Vanilla & 76.99 & 0.9967 & 0.3645 & 0.1290 \\
        Vanilla ensemble & 79.52 & 0.7368 & 0.3024 & 0.03856 \\ 
        \hline
        BN Adapt & 73.43 & 1.4502 & 0.4453 & 0.1878 \\
        Ensemble BN Adapt & 76.84 & 0.9444 & 0.3442 & 0.07855\\
        TENT (1 epoch) & 77.24 & 1.321 & 0.3896 & 0.1605 \\
        TENT (10 epochs) & 85.53 & 0.9554 & 0.2035 & 0.1166\\ % add regularized tent
        \hline
        TENT Ensemble (1 epoch) & 79.48 & 0.8869 & 0.3149 & 0.09254 \\
        TENT Ensemble (10 epochs) & 86.89 & 0.5784 & 0.2035 & 0.06384\\
        BACS (ours) (1 epoch) & 84.14 & 0.6024  & 0.2285 & 0.05595\\ 
        BACS (ours)  (10 epochs) & 86.32 & 0.5553 & 0.2094  & 0.05133 \\
        BACS - posterior (1 epoch) & 87.28 & 0.4214 & 0.1650 & 0.02603 \\ 
        BACS - posterior (10 epochs) & \textbf{93.03} & \textbf{0.2371} & \textbf{0.0965} & \textbf{0.02404 } \\
        \hline
    \end{tabular}
    \vspace{10pt}
    \caption{\textbf{SVHN to MNIST}: In this domain adaptation setting, we find adapting batch norm statistics alone hurts performance compared to the unadapated models, but methods utilizing entropy minimization are able to improve substantially in accuracy.}
    \vspace{-10pt}
    \label{tab:svhn_mnist}
\end{table}

We find that this severe distribution shift requires larger changes in parameters to effectively adapt, as seen by the substantial improvements between optimizing for one epoch and optimizing for 10. We also find that the approximate posterior term used in our method overly constrains the network during adaptation, as removing the regularizer from the method (denoted BACS-posterior in the table) results in the best performance in all metrics.
We note that all ensemble methods perform much better than non-ensembled methods in terms of calibration, emphasizing the benefits of marginalizing over different models for uncertainty estimation when adapting via entropy minimization.

We note that our method and the compared baselines are focused on improving results in robustness settings, and do not necessarily obtain state-of-the-art performance in common domain adaptation settings when compared to algorithms specifically designed for these problems. 
For example, \citet{Liang2020DoAdaptation} introduce a source-free domain adaptation algorithm (that also does not require access to the training data during adaptation) and report $99\%$ accuracy transferring from SVHN to MNIST, though results are not directly comparable due to architecture and training differences. 
In contrast to typical source-free domain adaptation algorithms, our algorithm also focuses on improving uncertainty estimation, does not require multiple epochs of optimization during adaptation, and is amenable to online evaluations where predictions need to be made before seeing the entirety of the test data.